\pdfoutput=1

\documentclass[11pt]{article}

\usepackage[final]{acl}

\usepackage{times}
\usepackage{latexsym}

\usepackage[T1]{fontenc}

\usepackage[utf8]{inputenc}

\usepackage{microtype}

\usepackage{inconsolata}

\usepackage{graphicx}

\usepackage{microtype}
\usepackage{url}
\usepackage{booktabs}

\usepackage{subfigure}
\usepackage{enumitem}
\newenvironment{itemize*}%
 {\leftmargini=20pt\begin{itemize}%
  \setlength{\itemsep}{3pt}%
  \setlength{\parskip}{0pt}%
  }%
 {\end{itemize}}
\newenvironment{enumerate*}%
 {\begin{enumerate}%
  \setlength{\itemsep}{0pt}%
  \setlength{\parskip}{0pt}}%
 {\end{enumerate}}

\usepackage{amsmath}
\usepackage{cleveref}
\usepackage{listings}
\usepackage{titlesec}
\usepackage{multirow}
\usepackage{makecell}

\usepackage{xcolor}
\usepackage{tcolorbox}

\definecolor{lightred}{RGB}{255,163,163}
\definecolor{deepred}{RGB}{146,0,0}

\usepackage{array}       
\usepackage{caption}     
\usepackage{amssymb}

\definecolor{midnightgreen}{rgb}{0.0, 0.29, 0.33}
\definecolor{deepgreen}{HTML}{0aa344}
\definecolor{deeppurple}{HTML}{7030a0}
\definecolor{deepblue}{HTML}{171d91}
\definecolor{brown}{HTML}{843c0c}
\definecolor{shadered}{HTML}{ffe5e5}
\definecolor{shadegreen}{HTML}{e5f7ed}

\newcommand{\red}{\textcolor{red}}
\newcommand{\green}{\textcolor{deepgreen}}

\NewDocumentCommand{\heng}
{ mO{} }{\textcolor{red}{\textsuperscript{\textit{Heng}}\textsf{\textbf{\small[#1]}}}}
\NewDocumentCommand{\cheng}
{ mO{} }{\textcolor{orange}{\textsuperscript{\textit{Cheng}}\textsf{\textbf{\small[#1]}}}}
\NewDocumentCommand{\avi}
{ mO{} }{\textcolor{blue}{\textsuperscript{\textit{Avi}}\textsf{\textbf{\small[#1]}}}}
\NewDocumentCommand{\xiusi}
{ mO{} }{\textcolor{purple}{\textsuperscript{\textit{Xiusi}}\textsf{\textbf{\small[#1]}}}}

\usepackage[normalem]{ulem}        

\title{SMART: Self-Aware Agent for Tool Overuse Mitigation}

\author{
Cheng Qian$^{1}$\thanks{\ Indicates equal contribution.},  Emre Can Acikgoz$^{1*}$, Hongru Wang$^{1}$\thanks{\ Mentorship}, Xiusi Chen$^{1}$, Avirup Sil$^{2}$,\\
\textbf{Dilek Hakkani-Tür$^{1}$, Gokhan Tur$^{1}$, Heng Ji$^{1}\footnotemark[2]$}\\
$^{1}$University of Illinois Urbana-Champaign, $^{2}$IBM Research AI\\
\texttt{\{chengq9, acikgoz2, hengji\}@illinois.edu}\\
}

\begin{document}
\maketitle
\begin{abstract}
Current Large Language Model (LLM) agents demonstrate strong reasoning and tool use capabilities, but often lack self-awareness, failing to balance these approaches effectively. This imbalance leads to \textbf{Tool Overuse}, where models unnecessarily rely on external tools for tasks solvable with parametric knowledge, increasing computational overhead.
Inspired by human metacognition, we introduce \textbf{SMART} (\uline{S}trategic \uline{M}odel-\uline{A}ware \uline{R}easoning with \uline{T}ools), a paradigm that enhances an agent’s self-awareness to optimize task handling and reduce tool overuse. 
To support this paradigm, we introduce \textbf{SMART-ER}, a dataset spanning three domains, where reasoning alternates between parametric knowledge and tool-dependent steps, with each step enriched by rationales explaining when tools are necessary.
Through supervised training, we develop \textbf{SMARTAgent}, a family of models that dynamically balance parametric knowledge and tool use.
Evaluations show that SMARTAgent reduces tool use by 24\% while improving performance by over 37\%, enabling 7B-scale models to match its 70B counterpart and GPT-4o. Additionally, SMARTAgent generalizes to out-of-distribution test data like GSM8K and MINTQA, maintaining accuracy with just one-fifth the tool calls.
These highlight the potential of strategic tool use to enhance reasoning, mitigate overuse, and bridge the gap between model size and performance, advancing intelligent and resource-efficient agent designs.
All the data and codes are released\footnote{\ \begin{scriptsize}\url{https://github.com/qiancheng0/Open-SMARTAgent}\end{scriptsize}}.
\end{abstract}

\section{Introduction}

Recent advancements in Large Language Models (LLMs) \citep{ouyang2022training,team2023gemini,dubey2024llama} have led to remarkable improvements in reasoning capabilities, driving progress in diverse domains such as coherent text composition~\citep{wei2023chainofthought}, code generation~\citep{gao2023pal, Opendevin2024, SWEGym2024}, complex logical deduction~\citep{yao2023react, yao2024tree}, and nuanced natural language understanding~\citep{wang2023interactive, yu2024natural, wu2024individual}.
However, challenges remain, such as the inability to handle real-time information~\cite{yu2024information}, model real-world challenges~\citep{qian2025modelingagent}, provide accurate mathematical results~\citep{lu2022survey}, and fully comprehend human intentions~\cite{qian2024tell}. These limitations highlight the need for LLMs to leverage external tools~\citep{schick2023toolformer, qin2023tool, yuan2024craft, qian2024escapebench}, enabling them to function as agents capable of assisting users in diverse tasks~\citep{qin2023toolllm, xi2023rise}. Effective tool use and reasoning are thus complementary, each enhancing the other to overcome current shortcomings.


Therefore, in problem-solving, a language agent often combines reasoning with tool use, following a ReACT-style approach~\citep{yao2023react}, where the model alternates between thought processes and actions to derive solutions. This enables the core agent to apply its parametric knowledge to advance task-solving while using external tools to address its limitations. However, this interplay raises a critical question: \textbf{when should the agent rely on external tools versus its own knowledge?} 


\begin{figure}[!t]
    \centering
    \subfigure{\includegraphics[width=\linewidth]{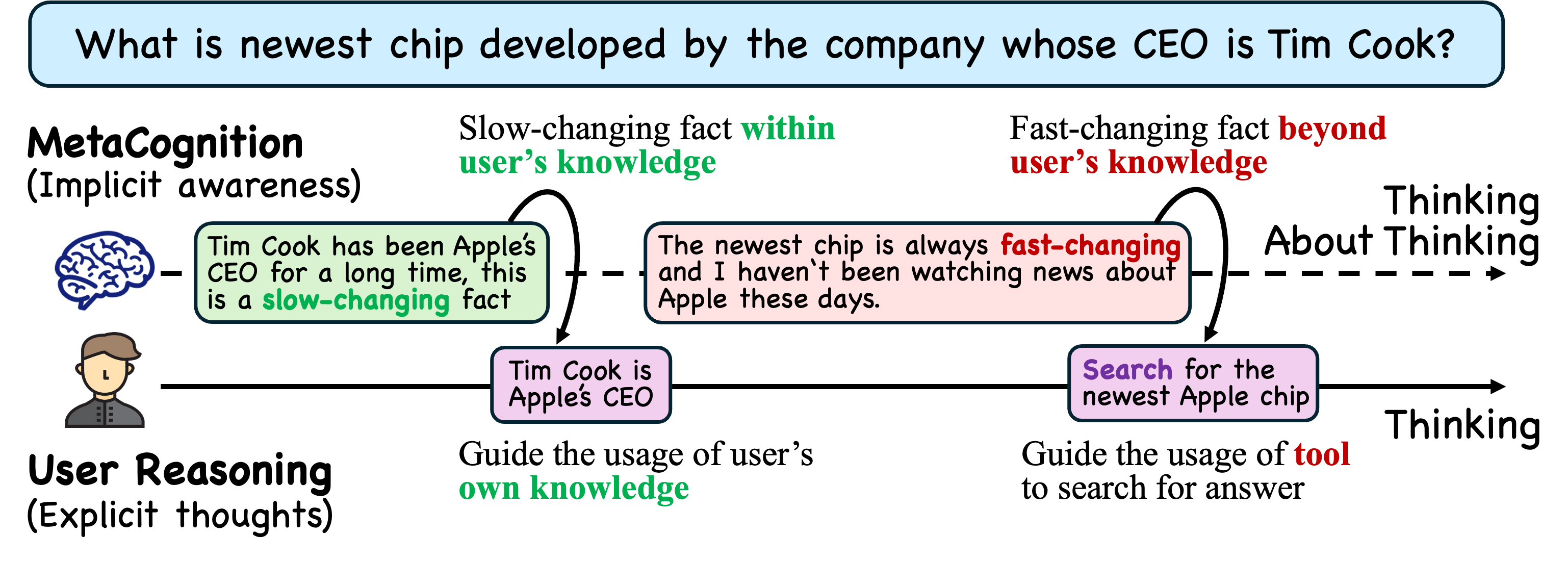}}
    \vspace{-5mm}
    \caption{An illustration of human metacognition: The user recalls Tim Cook’s role from \textbf{prior knowledge} (a \textit{slow-changing fact}), but uses \textbf{online search} to find the latest chip info (a \textit{fast-changing fact}).}
    \vspace{-5mm}
    \label{fig:intro_metacognition}
\end{figure}

To investigate this, we first conduct a preliminary study on both LLMs and LM-driven agent systems to assess their ability to dynamically and effectively switch between external tool use and parametric knowledge-driven reasoning. Our empirical results reveal a consistent bias, with LLMs unnecessarily invoking tools over 30\% of the time, and agent systems exhibiting similar behavior even when their parametric knowledge alone would suffice. We identify this phenomenon as \textbf{Tool Overuse}, which arises from the model's inability to recognize when its internal knowledge is sufficient. This not only leads to unnecessary resource consumption but can also confuse the model, ultimately degrading performance. This observation highlights the need for better calibration of an agent’s self-awareness, ensuring it can discern when to rely on tools versus its own knowledge. Striking this balance is crucial for enhancing efficiency, scalability, and user experience as LM-driven agents are increasingly deployed in real-world applications.



To address this challenge, we propose \textbf{SMART} (\uline{S}trategic \uline{M}odel-\uline{A}ware \uline{R}easoning with \uline{T}ools), which draws inspiration from human decision-making to calibrate self-awareness in agent models for effective tool use and reasoning. In \textbf{Metacognitive Theory}~\citep{schraw1995metacognitive}, psychology highlights humans’ awareness of their thought processes, including when to apply specific problem-solving strategies~\citep{livingston2003metacognition}. As \Cref{fig:intro_metacognition} illustrates, this implicit heuristic allows dynamic balancing between external strategies and internal knowledge~\citep{Minsky1986-MINTSO}. Similarly, agents need metacognition to optimize tool usage. By aligning the model’s subjective perception with its knowledge boundary, we enable agents to make more informed decisions on when to rely on external tools or internal knowledge.

We adopt a data-driven approach to calibrate model decision-making by constructing \textbf{SMART-ER} (SMART-\uline{E}nhanced \uline{R}easoning), a dataset spanning three domains—Math, Time, and Intention. It addresses key LLM limitations, including computational accuracy~\citep{hendrycks2021measuring}, outdated knowledge~\citep{vu2023freshllms}, and user preference awareness~\citep{qian2024tell}. Specifically, each question in SMART-ER combines sub-questions the model handles well (e.g., simple arithmetic, static facts, commonsense) with those it struggles with (e.g., complex math, dynamic facts, user-specific intentions).
We break down each question into reasoning steps, categorizing them as either parametric knowledge-driven or tool-dependent. For parametric steps, we provide reasoning based on internal knowledge. For tool-dependent steps, we map them to appropriate tools, execute them, and integrate the results into the reasoning process.
Finally, inspired by metacognitive heuristics, we refine each step with explicit justifications, clarifying when parametric knowledge suffices or external tools are needed. By transforming implicit decision-making heuristics into explicit language-based reasoning, we guide the model to develop calibrated awareness of its knowledge boundaries.

Leveraging SMART-ER, we develop \textbf{SMARTAgent}, a family of agent models designed to dynamically balance reasoning between parametric knowledge and external tools. Empirical results show that SMARTAgent reduces tool use by 24\% while improving overall performance by over 37\%, effectively mitigating tool overuse. Notably, it enables 7B-scale models to match the performance of GPT-4 and 70B models, bridging the gap between model size and capability. Additionally, SMARTAgent efficiently handles out-of-distribution (OOD) tasks, requiring only one-fifth the number of tool calls while preserving accuracy. Finally, analysis of SMARTAgent’s confidence through logits reveals more certain reasoning-tool-switching decisions, further validating our approach in calibrating the agent’s self-awareness. In summary:
\begin{itemize}[topsep=2pt, partopsep=-5pt, leftmargin=8pt, itemsep=-4.5pt]
\item We identify and define the issue of \textbf{Tool Overuse}, emphasizing that strategically balancing the complementary strengths of knowledge-driven reasoning and external tool calls can mitigate this problem in both LLMs and agent systems.
\item We introduce \textbf{SMART-ER}, a multi-domain dataset designed to address key limitations of agent models by integrating metacognitive heuristics to better help them recognize and adapt to their knowledge boundaries.
\item We develop \textbf{SMARTAgent}, a family of agents that intelligently balances parametric reasoning and tool use, achieving improved performance, reduced tool overuse, and more confident decision-making in tool utilization.
\end{itemize}

\section{Related Work}



\paragraph{LM Knowledge Boundary.}
Recent studies highlight that while LMs excel at standard tasks, they struggle to recognize and acknowledge the limits of their knowledge~\cite{yin2023large, qian2023merge, kadavath2022language}. To address this gap, the concept of knowledge boundary has been introduced to define the limits of knowledge in LLMs~\cite{li2024knowledge, amayuelas2023knowledge}.
Building on this, some research evaluates LMs' self-awareness of their knowledge boundary through verbal probing~\citep{kadavath2022language} and fine-grained benchmarks~\citep{yin2024benchmarking}, enabling LMs to determine whether a question is answerable.
Other work focuses on mitigating hallucinations arising from the model’s unawareness of its limits through data augmentation~\cite{chen2023gotta,chen-etal-2024-minprompt}, retrieval augmentation~\cite{ren2023investigating}, and confidence calibration~\citep{xue2024ualign}. Additionally, \citet{chen2024teaching} and \citet{zhang2024r} trained LLMs to express their knowledge boundaries, enabling them to answer known questions and admit ignorance for unknown ones.
Recently, reinforcement learning has been increasingly explored as a means to help models recognize their knowledge boundaries to guide more efficient decisions~\citep{qian2025toolrl, wang2025otc}.
Our work aligns with these studies and focuses on enhancing agents' awareness for wiser tool use.

\paragraph{LM Tool Use.}
Integrating tool use into LLMs has gained significant attention as a way to complement parametric knowledge and enhance decision-making~\cite{qin2023tool, qu2025tool}. Some research focuses on enabling LLMs to access external tools to overcome knowledge limitations~\citep{qin2023toolllm, qian2024toolink}, including up-to-date information~\citep{vu2023freshllms, wang2024uniretriever} and domain-specific expertise~\citep{ling2023domain, wang2024appbench}. Others explore tool creation~\citep{qian2023creator, cai2023large} and external module integration~\citep{qian2024investigate} to improve tool learning robustness. Despite these, a key challenge lies in evaluating and enhancing LLMs’ ability to determine when and which tools to use. Benchmarks like MetaTool~\cite{huang2023metatool} and WTU-EVAL~\cite{ning2024wtu} highlight LLMs' struggles with unnecessary or incorrect tool usage, while dynamic frameworks~\cite{wang2024selfdc, shen2024smartcal} propose adaptively invoking tools based on internal uncertainty thresholds. Unlike prior works, SMART rigorously defines \textit{tool overuse} and addresses it by optimizing the balance of internal knowledge and tool use.

\section{Preliminaries}


\begin{figure}[!t]
    \centering
    \subfigure{\includegraphics[width=\linewidth]{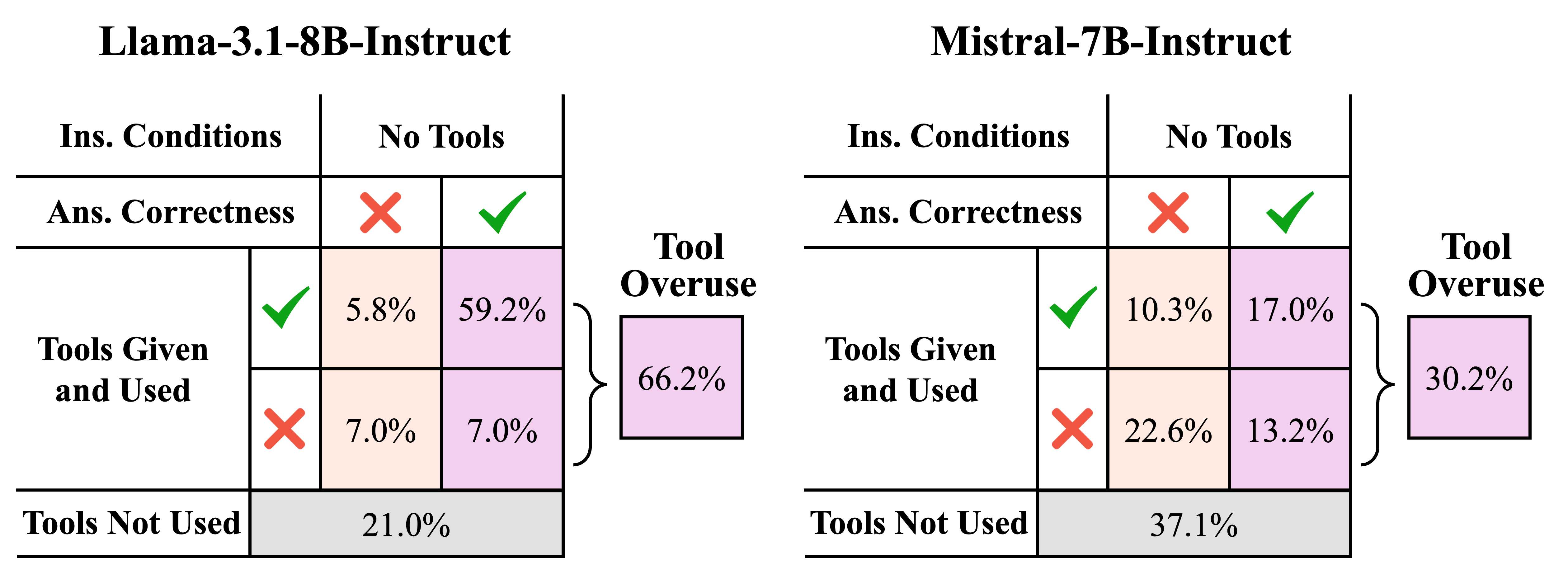}}
    \vspace{-5mm}
    \caption{Statistics on Llama and Mistral's tool overuse.}
    \label{fig:preliminary_model}
\end{figure}



To investigate how models decide between invoking tools and relying on their own knowledge, we conduct a preliminary study on both LLMs and LM-driven agent systems. Our findings reveal both LLMs and agent systems' strong tendency for excessive tool use, which we define as \textbf{Tool Overuse}, leading to unnecessary resource overhead.

\vspace{2mm}
\noindent \textbf{Definition of Tool Overuse.} \label{sec:tool_overuse}
Tool overuse refers to the excessive reliance on external tools when an agent model could have successfully completed the task using its parametric knowledge alone.
Formally, let \( Q \) be the total set of questions, and let \( P \) be the subset of questions that the model can correctly answer without using any tools. The model's intrinsic reasoning capability is then given by \( \alpha = \frac{|P|}{|Q|} \). Now, suppose that when provided with access to tools, the model chooses to invoke at least one tool on a fraction \( \beta \) of these questions in \( P \). The \textbf{Tool Overuse Rate} is then defined as:
\vspace{-3mm}
\[
\mathcal{O} = \alpha \cdot \beta
\]

\vspace{-3mm} \noindent which quantifies the proportion of all questions where tool use is unnecessary, highlighting inefficiencies in the model's decision-making process.

\begin{table}[!t]
    \centering
    \subfigure{\includegraphics[width=\linewidth]{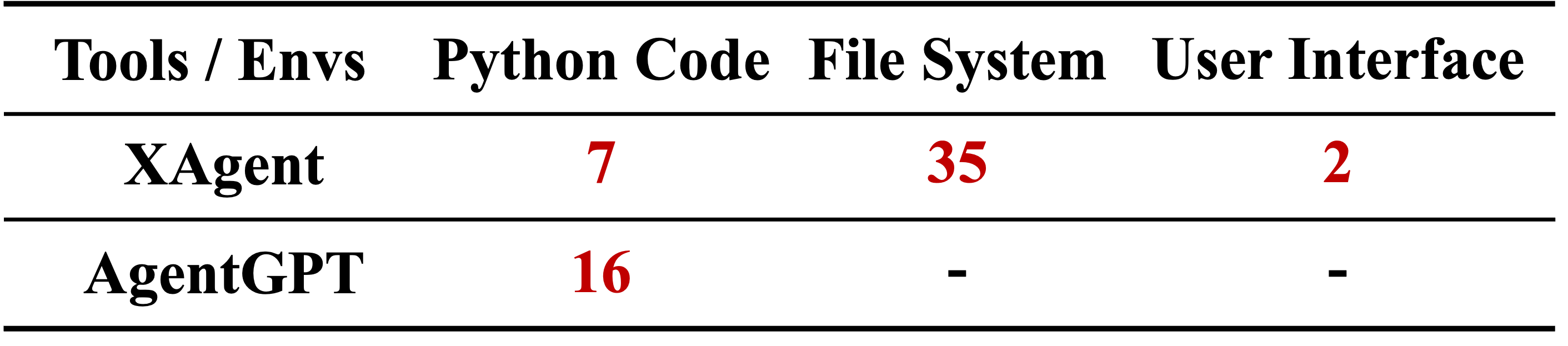}}
    \vspace{-3mm}
    \caption{Statistics on XAgent and AgentGPT's tool overuse. Both agents invoke tools multiple times across 50 samples, despite \textit{ideally} requiring \textit{zero} tool usage.}
    \label{tab:preliminary_agent_table}
\end{table}

\vspace{2mm}
\label{sec:preliminary_llm}
\noindent \textbf{Experiments on LLMs.}
We first experiment with Llama-3.1-8B~\citep{dubey2024llama} and Mistral-7B~\citep{jiang2023mistral} on the GSM8K test set~\citep{cobbe2021training}. Each test question is presented under two conditions: i) the model reasons through the question normally and provides a final answer without using tools, and ii) the model has access to tools and independently decides whether to use them (see \Cref{apdx:prelim_model}). The statistics in \Cref{fig:preliminary_model} reveal two key insights. First, both models exhibit significant tool overuse, with Llama's rate exceeding 50\%. Second, in some cases, tool use leads to incorrect answers, even for questions the model could have solved correctly without external assistance. This highlights how excessive reliance on tools can introduce unnecessary complexity and degrade performance.


\begin{figure}[!t]
    \centering
    \subfigure{\includegraphics[width=\linewidth]{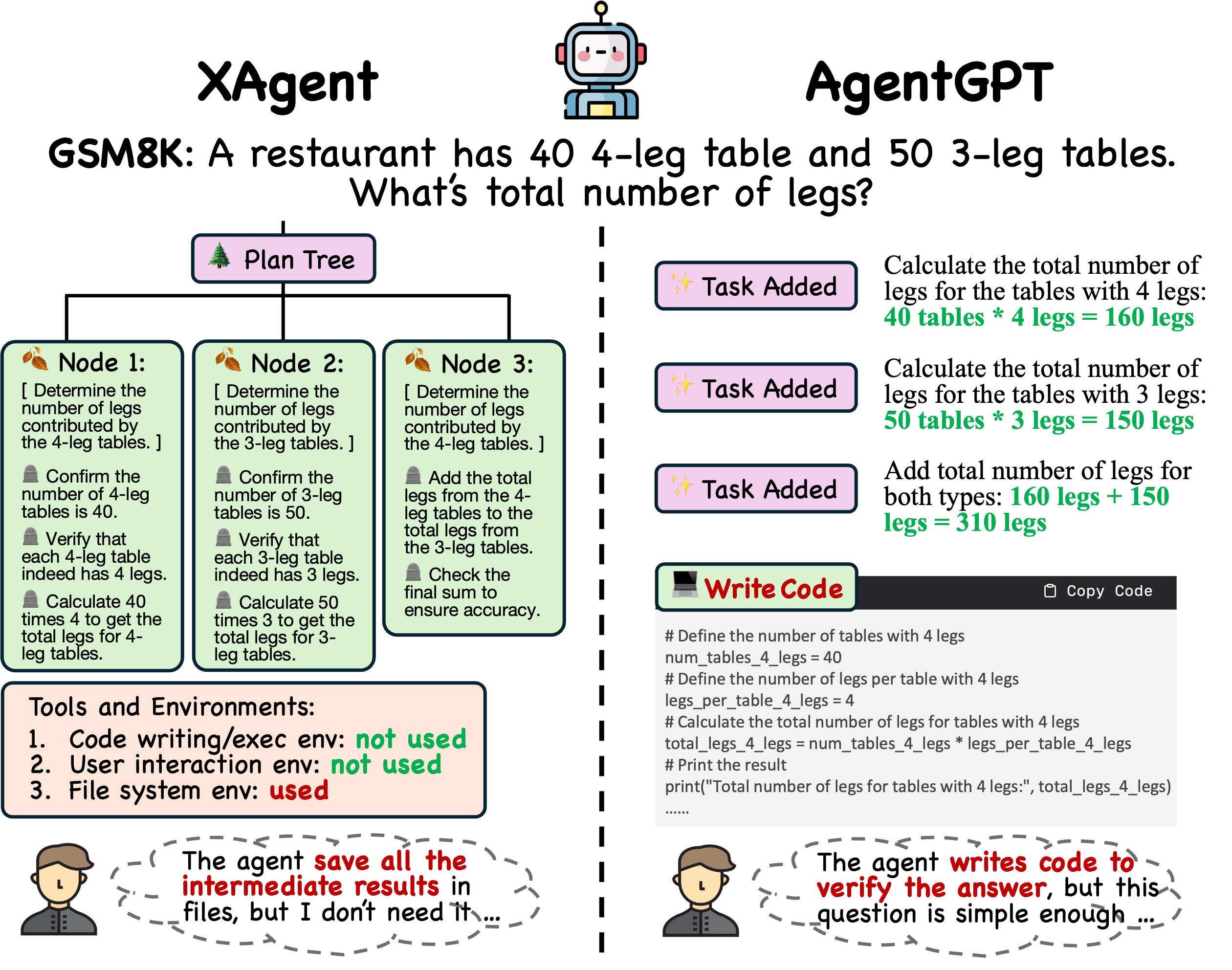}}
    \vspace{-5mm}
    \caption{Example cases on XAgent and AgentGPT's tool overuse.}
    \label{fig:preliminary_agent}
\end{figure}

\vspace{2mm}
\noindent \textbf{Experiments on LM-driven Agents.} 
In addition to LLMs, we also experiment with two agent systems: XAgent~\citep{xagent2023} and AgentGPT~\citep{2024agentgpt}, both designed for complex problem-solving and driven by closed-source GPT models. We sampled 50 queries from the GSM8K test set that can be answered correctly without tools (see \Cref{apdx:prelim_agent}) and instructed the models to use tools only when necessary. The results in \Cref{tab:preliminary_agent_table} show that, despite being equipped with various tools, both agent systems still tend to use them unnecessarily, significantly slowing down problem-solving (about 10x slower than using GPT alone). We further provide a case study in \Cref{apdx:prelim_agent} highlighting issues such as XAgent redundantly saving results to files and AgentGPT unnecessarily invoking a code-writing tool after generating an answer. These observations underscore the need to address our core research question: \textbf{How can we calibrate agent models to balance tool use and parametric reasoning, mitigating tool overuse while preserving utility?}

\section{Method}

\begin{figure}[!t]
    \centering
    \subfigure{\includegraphics[width=\linewidth]{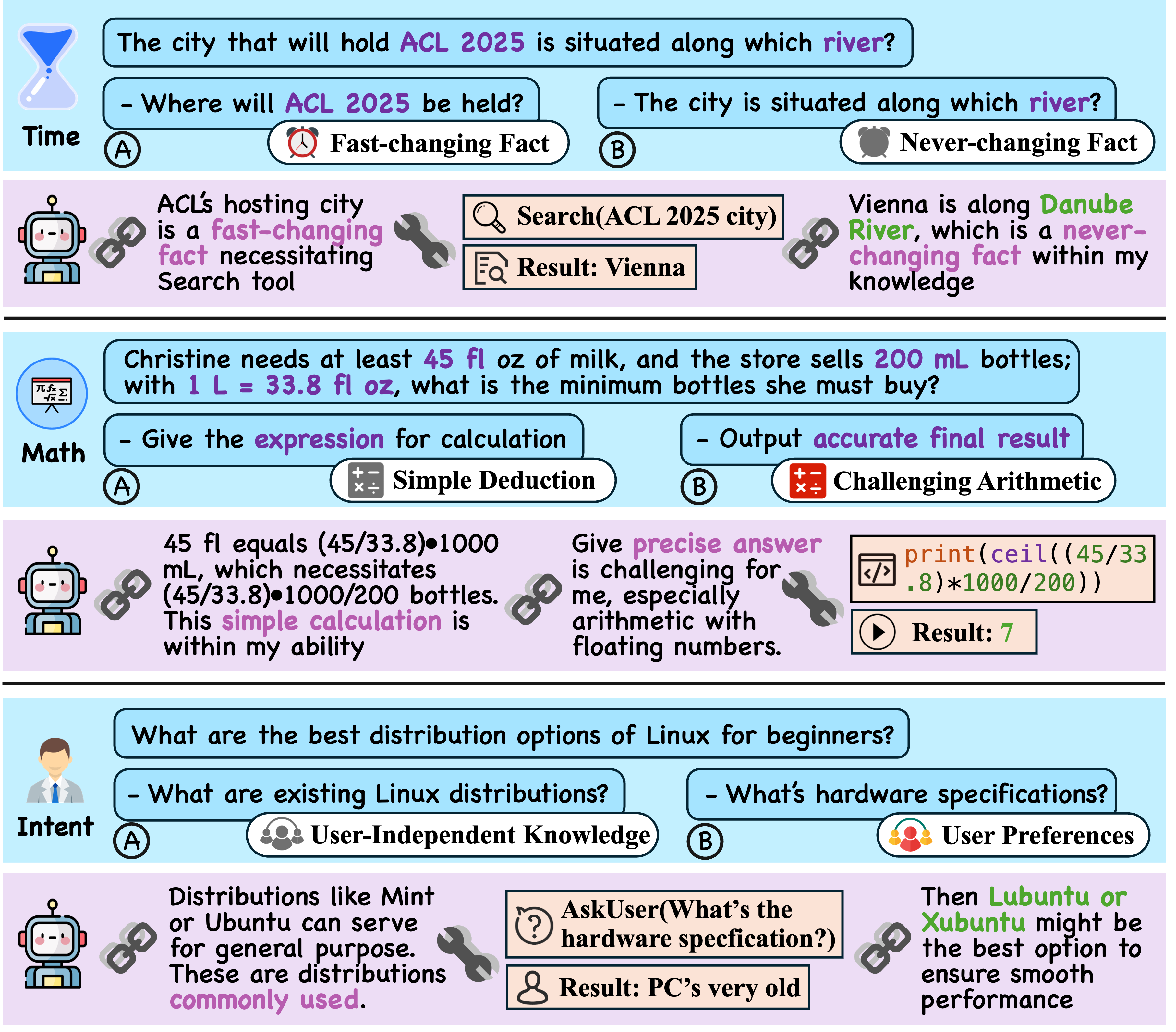}}
    \vspace{-5mm}
    \caption{Three example queries and their reasoning chains from each domain. The inherent compositionality of a query naturally divides reasoning into knowledge-driven steps and tool-reliant steps.}
    \label{fig:intro_case}
\end{figure}

To address the challenge of tool overuse, we draw inspiration from how humans balance internal knowledge and external tools. Metacognitive theory~\citep{schraw1995metacognitive} suggests that human decision-making relies on an implicit awareness of knowledge boundaries, enabling strategic, step-by-step problem-solving~\citep{livingston2003metacognition}. Inspired by this, we aim to equip agent models with a similar capability—calibrating their metacognition to optimize reasoning and tool use.



To address this, we propose \textbf{SMART}, a data-driven approach that enhances self-awareness in agent models. While LLMs acquire broad knowledge from large-scale corpora~\citep{wang2022generalizing}, they are not explicitly trained to recognize their own strengths and limitations. To bridge this gap, we introduce \textbf{SMART-ER}, the first dataset contrasting areas where models excel versus struggle. Covering three domains with 3K+ questions and structured reasoning chains, SMART-ER helps agents strategically decide when to rely on internal knowledge or external tools.  




\begin{figure*}[!t]
    \centering
    \subfigure{\includegraphics[width=\linewidth]{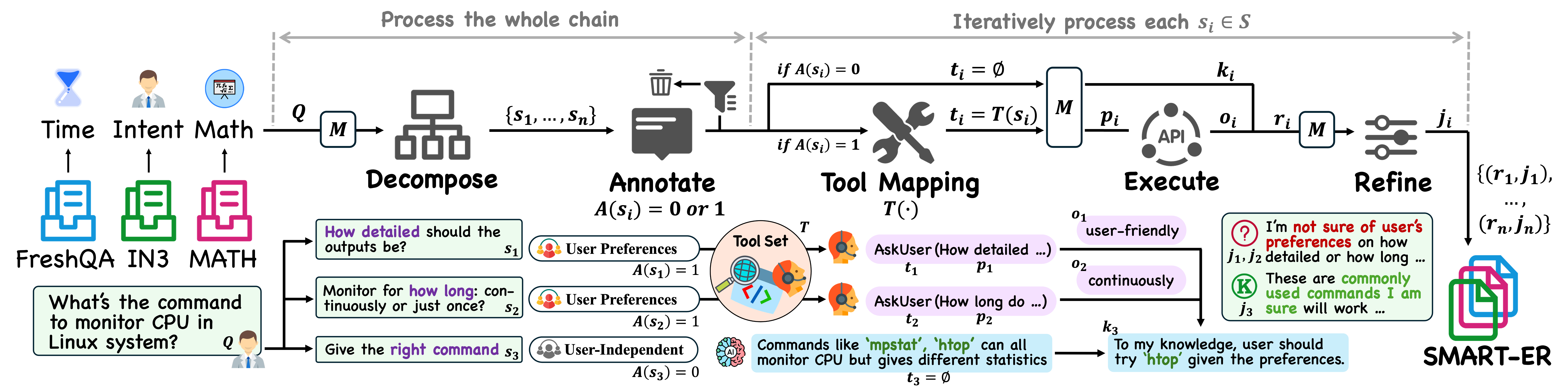}}
    \vspace{-5mm}
    \caption{The data pipeline to get SMART-ER. We divide the whole pipeline into several stages for better control and quality of the generated reasoning chain.}
    \label{fig:pipeline}
\end{figure*}

\subsection{Data Collection}
To train agents to strategically balance parametric knowledge and external tools within a single reasoning chain, questions must be compositional—blending aspects the model excels at with those it struggles with. Building on prior studies~\citep{hendrycks2021measuring, vu2023freshllms, qian2024tell}, we identify three key limitations in LMs: i) \textit{math reasoning}, where models struggle with complex computations requiring precise answers; ii) \textit{temporal knowledge}, as LMs lack access to up-to-date facts beyond their training cutoff; and iii) \textbf{user intent understanding}, where implicit preferences cannot be inferred without direct queries.
All these challenges necessitate a smarter integration of external tools with the model’s reasoning ability. Building on this insight, we construct data of three domains: 
\begin{itemize}[topsep=2pt, partopsep=-5pt, leftmargin=8pt, itemsep=-4.5pt]
\item \textbf{Math}: Adapted from MATH \citep{hendrycks2021measuring}, each query incorporates both challenging math deductions and simple arithmetic to contrast reasoning capabilities.
\item \textbf{Time}: Adapted from FreshQA \citep{vu2023freshllms}, each query ensures a mix of fast-changing and slow-changing factual knowledge.
\item \textbf{Intention}: Adapted from Intention-in-Interaction (IN3) \citep{qian2024tell}, each query requires explicit user intent while remaining solvable within the model’s capabilities.
\end{itemize}
This compositional approach helps models calibrate their decision-making by distinguishing when to rely on external tools versus when internal knowledge is sufficient. To illustrate this, we present three example queries from each domain in \Cref{fig:intro_case}. For details on the question selection and adaptation process, please refer to \Cref{apdx:data_slection}.

\begin{table}[!t]
    \centering
    \subfigure{\includegraphics[width=\linewidth]{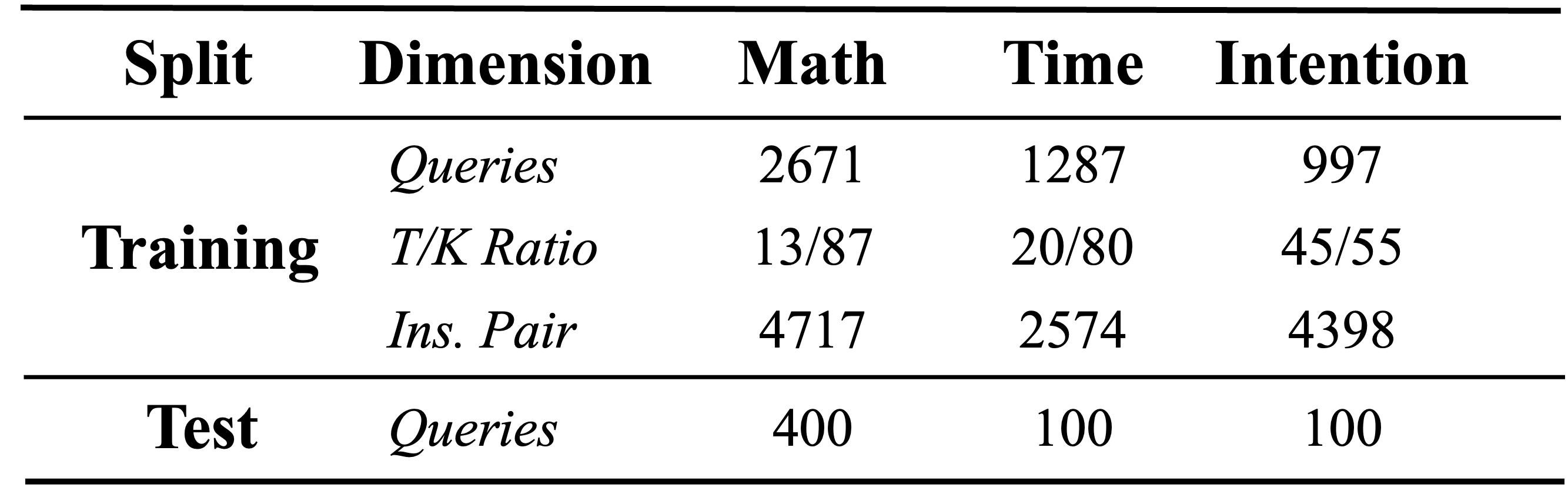}}
    \vspace{-3mm}
    \caption{Statistics for SMART-ER. \textit{T/K Ratio} denotes the ratio of tool-reliant to knowledge-driven steps.}
    \label{tab:data_statistic}
\end{table}

\subsection{Reasoning Chain Construction}
As shown in \Cref{fig:pipeline}, each query \( Q \) is decomposed into a structured reasoning plan with \( n \) subgoals, \( S = \{s_1, s_2, \dots, s_n\} \). This decomposition is enabled by the compositional nature of our queries and is empirically achieved using GPT-4o, an auxiliary model in our pipeline, later denoted as \( M \).
Next, for each \( s_i \), we determine whether it requires tool use (\( A(s_i) = 1 \)) or can be resolved with parametric knowledge alone (\( A(s_i) = 0 \)). Using ground truth from existing source data as heuristics, we guide \( M \) to annotate each subgoal. During this process, we also discard those queries where all subgoals rely exclusively on either tools or parametric knowledge.
After annotating the entire chain, we process each subgoal iteratively, starting from \( s_1 \). For each subgoal \( s_i \) where \( A(s_i) = 1 \), we assign an appropriate tool \( t_i \) from a predefined tool set using a mapping function \( T(\cdot) \):
\vspace{-1mm}
\[
t_i =
\begin{cases} 
T(s_i), & \text{if } A(s_i) = 1 \\ 
\varnothing, & \text{otherwise} 
\end{cases}
\]

\vspace{-1mm} \noindent where \( t_i = \varnothing \) indicates the model relies solely on its parametric knowledge for reasoning. Empirically, our tool set consists of \textit{Code}, \textit{Search}, and \textit{AskUser}, covering all designed domains.

Next, we proceed with the reasoning process using \( M \). If \( A(s_i) = 0 \), \( M \) reasons over \( s_i \), producing a reasoning step \( k_i \) based on its parametric knowledge. Otherwise, we prompt \( M \) to generate the necessary parameters \( p_i \) for tool invocation, retrieving the tool output \( o_i \). The resulting outcome for each step is formulated as:
\vspace{-1mm}
\[
r_i =
\begin{cases} 
(\ p_i = M(s_i),\ o_i = t_i(p_i)\ ), & \text{if } A(s_i) = 1 \\ 
(\ k_i = M(s_i)\ ), & \text{otherwise} 
\end{cases}
\]

\vspace{-1mm} \noindent where \( t_i(\cdot) \) represents the invocation of tool \( t_i \). The iterative process also enables \( M \) to incorporate information from prior steps and tool outputs when processing subsequent subgoals, ensuring a coherent and context-aware reasoning flow.

Inspired by metacognitive heuristics that implicitly guide human reasoning, we refine the reasoning chain \( r_i \) by explicitly incorporating justifications for whether parametric knowledge suffices or external tool use is necessary.
Specifically, we prompt \( M \) to generate a justification \( j_i = M(s_i, A(s_i)) \), conditioned on the subgoal \( s_i \) and its annotation \( A(s_i) \). This approach emulates human metacognition by transforming implicit heuristics into explicit natural language explanations, thus enhancing interpretability. Similar to Chain-of-Thought~\citep{wei2022chain} leverages the cumulative probability nature of autoregressive models to guide reasoning, \( j_i \) helps the model calibrate its decision-making, improving its ability to strategically balance internal knowledge and external tools.

Finally, by integrating all subgoals, we obtain the complete reasoning chain \( R = \{(r_1, j_1), \dots, (r_n, j_n)\} \) for query \( Q \), where each step \( r_i \) is either \( (k_i) \), indicating a parametric knowledge-driven step, or \( (p_i, o_i) \), representing a tool-reliant step. Our method dynamically integrates these steps, ensuring an adaptive balance between internal reasoning and external tool use. To ensure quality, we conduct human supervision on 5\% of the data for each step involving \( M \), achieving a pass rate of over 95\%. Please refer to \Cref{apdx:chain_construction} for details.




\subsection{Agent Training Implementation}
We partition SMART-ER into training and test splits with statistics in \Cref{tab:data_statistic}. For each \((Q, R')\) in the training set, we generate multiple input-output pairs for instruction tuning. The input comprises \(\{Q, (r_1, j_1), \dots, (r_{x_i}, j_{x_i})\}\), while the output consists of \(\{(r_{{x_i}+1}, j_{{x_i}+1}), \dots, (r_{x_{i+1}}, j_{x_{i+1}})\}\},\) where \( x_i \) indexes the tool-reliant steps. This setup ensures iterative reasoning, allowing the agent to leverage prior steps until the next tool invocation or final solution. The number of input-output pairs per \((Q, R')\) also equals the number of tool-reliant steps, facilitating interactive inference.


Using these instruction pairs, we finetune the Llama-3.1 8B and 70B instruct models~\citep{dubey2024llama} as well as the Mistral 7B, Nemo(12B) and Small(24B) instruct models~\citep{jiang2023mistral}, adapting them into a family of \textbf{SMARTAgent}. These agent models enable interactive tool use, recognizes its own limitations, and balances tool reliance with parametric knowledge-driven reasoning to prevent tool overuse. See \Cref{apdx:training} for training details and hyper-parameters.

\begin{table*}[t]
    \centering
    \vspace{-4mm}
    \setlength\tabcolsep{2pt}
    \setlength\extrarowheight{2pt}
    
    \resizebox{1.0\linewidth}{!}{
    \begin{tabular}{l @{\hskip 14pt} l c c @{\hskip 14pt} c c @{\hskip 14pt} c c c}
    
        \toprule
        
        \multirow{2}{*}{\textbf{Method}} & \multirow{2}{*}{\textbf{Model}} & \multicolumn{2}{c}{\textbf{Math} (MATH)} & \multicolumn{2}{c}{\textbf{Time} (FreshQA)} & \multicolumn{3}{c}{\textbf{Intention} (Intention-in-Interaction)} \\

        \cmidrule(lr){3-4} \cmidrule(lr){5-6} \cmidrule(lr){7-9}
        
        ~ & ~ & \textbf{\makecell{Tool Used$^\downarrow$\\(\textit{Times})}} & \hskip 8pt \textbf{\makecell{Accuracy$^\uparrow$\\(\%)}} & \textbf{\makecell{Tool Used$^\downarrow$\\(\textit{Times})}} & \hskip 8pt \textbf{\makecell{Accuracy$^\uparrow$\\(\%)}} & \textbf{\makecell{Tool Used$^\downarrow$\\(\textit{Times})}}  & \hskip 8pt \textbf{\makecell{Missing Details Recovery$^\uparrow$\\(Lv3 / Lv2, \%)}} & \textbf{\makecell{Summarized Intention \\Coverage$^\uparrow$ (\%)}}  \\

        \addlinespace[2pt]
        \midrule
        \addlinespace[2pt]
        \multicolumn{9}{c}{ \textit{Open-Source}} \\ 
        \midrule

        \multirow{2}{*}{\makecell[l]{Normal\\Reasoning Trained}} & \textit{Mistral-7B} & 0.00 & 17.00 & 0.00 & 48.00 & 0.00 & 41.86 / 43.84 & - \\
        ~ & \textit{Llama-3.1-8B} & 0.00 & 41.00 & 0.00 & 48.00 & 0.00 & 38.37 / 42.49 & - \\

        \addlinespace[2pt]
        \midrule
        \addlinespace[2pt]

        \multirow{5}{*}{\makecell[l]{Base Model\\Reasoning Prompt}} & \textit{Mistral-7B} & 0.00 & 17.25 & 0.00 & 29.00 & 0.00 & 37.21 / 33.06 & - \\ 
        ~ & \textit{Llama-3.1-8B} & 0.00 & 53.00 & 0.00 & 26.00 & 0.00 & 40.70 / 25.76 & - \\
        ~ & \textit{Mistral-Nemo(12B)} & 0.00 & 47.00 & 0.00 & 33.00 & 0.00 & 44.19 / 28.37 & - \\
        ~ & \textit{Mistral-Small(24B)} & 0.00 & 72.25 & 0.00 & 34.00 & 0.00 & 41.86 / 31.82 & - \\ 
        ~ & \textit{Llama-3.1-70B} & 0.00 & 70.00 & 0.00 & 36.00 & 0.00 & 41.86 / 29.24 & - \\

        \addlinespace[2pt]
        \midrule
        \addlinespace[2pt]
        
        \multirow{5}{*}{\makecell[l]{Base Model\\Tool Prompt}} & \textit{Mistral-7B} & 3.90 & 13.25 & 1.67 & 49.00 & 3.80 & 48.84 / 21.70 & 63.04 \\ 
        ~ & \textit{Llama-3.1-8B} & 1.93 & 51.00 & 2.05 & 56.00 & 3.77 & 54.76 / 25.90 & 70.20 \\ 
        ~ & \textit{Mistral-Nemo(12B)} & 2.35 & 46.00 & 1.19 & 59.00 & 1.80 & 31.35 / 5.82 & 59.27 \\
        ~ & \textit{Mistral-Small(24B)} & 1.55 & 76.00 & 1.73 & 62.00 &  2.52 & 45.74 / 33.62 & 78.20 \\
        ~ & \textit{Llama-3.1-70B} & 3.53 & 67.50 & 2.08 & 63.00 & 2.71 & 45.74 / 35.96 & 61.68 \\ 

        \addlinespace[2pt]
        \midrule
        \addlinespace[2pt]

        \multirow{6}{*}{\textbf{SMARTAgent}} & \textit{Mistral-7B} & 0.60$_{\green{\downarrow 3.30}}$ & 22.75$_{\green{\uparrow 5.50}}$ & 1.00$_{\green{\downarrow 0.67}}$ & 64.00$_{\green{\uparrow 15.00}}$ & 3.60$_{\green{\downarrow 0.20}}$ & 74.42$_{\green{\uparrow 25.58}}$ / 65.44$_{\green{\uparrow 21.60}}$ & 81.76$_{\green{\uparrow 18.72}}$ \\  
        ~ & \textit{Llama-3.1-8B} & 0.88$_{\green{\downarrow 1.05}}$ & 54.75$_{\green{\uparrow 1.75}}$ & 1.05$_{\green{\downarrow 1.00}}$ & 67.00$_{\green{\uparrow 11.00}}$ & 3.80$_{\red{\uparrow 0.03}}$ & \textbf{81.40}$_{\green{\uparrow 26.64}}$ / 67.41$_{\green{\uparrow 24.92}}$ & 78.28$_{\green{\uparrow 8.08}}$ \\
        ~ & \textit{Mistral-Nemo(12B)} & 0.82$_{\green{\downarrow 1.53}}$ & 49.50$_{\green{\uparrow 2.50}}$ & 1.00$_{\green{\downarrow 0.19}}$ & \textbf{70.00}$_{\green{\uparrow 11.00}}$ & 3.34$_{\red{\uparrow 1.54}}$ & 77.91$_{\green{\uparrow 33.72}}$ / 62.15$_{\green{\uparrow 33.78}}$ & 82.30$_{\green{\uparrow 23.03}}$ \\
        ~ & \textit{Mistral-Small(24B)} & 0.79$_{\green{\downarrow 0.76}}$ & 69.75$_{\red{\downarrow 6.25}}$ & 1.00$_{\green{\downarrow 0.73}}$ & 66.00$_{\green{\uparrow 4.00}}$ & 3.89$_{\red{\uparrow 1.37}}$ & 74.42$_{\green{\uparrow 28.68}}$ / \textbf{68.87}$_{\green{\uparrow 35.25}}$ & 84.99$_{\green{\uparrow 6.79}}$ \\ 
        ~ & \textit{Llama-3.1-70B} & 0.94$_{\green{\downarrow 2.59}}$ & \textbf{72.50}$_{\green{\uparrow 2.50}}$ & 1.01$_{\green{\downarrow 1.07}}$ & 66.00$_{\green{\uparrow 3.00}}$ & 3.51$_{\red{\uparrow 0.80}}$ & 68.60$_{\green{\uparrow 22.86}}$ / 58.15$_{\green{\uparrow 22.19}}$ & \textbf{86.09}$_{\green{\uparrow 24.41}}$ \\
        
        \addlinespace[2pt]
        \cmidrule(lr){2-9}
        
        ~ & \multicolumn{3}{r}{Tool Used Macro-Average Decrease (\%)} & \textbf{24.00} & \multicolumn{3}{r}{Performance Macro-Average Increase (\%)} & \textbf{37.10} \\
        
        \addlinespace[2pt]
        \midrule
        \addlinespace[2pt]
        \multicolumn{9}{c}{ \textit{Closed-Source}} \\ 
        \midrule

        \multirow{2}{*}{\makecell[l]{Base Model\\Reasoning Prompt}} & \textit{GPT-4o-mini} & 0.00 & 73.00 & 0.00 & 44.00 & 0.00 & 45.35 / \textbf{32.41} & - \\ 
        ~ & \textit{GPT-4o} & 0.00 & \textbf{79.50} & 0.00 & 47.00 & 0.00 & 38.37 / 28.54 & - \\ 

        \addlinespace[2pt]
        \midrule
        \addlinespace[2pt]
        
        \multirow{2}{*}{\makecell[l]{Base Model\\Tool Prompt}} & \textit{GPT-4o-mini} & 2.55 & 54.50 & 1.06 & 56.00 & 1.91 & \textbf{50.00} / 26.90 & 76.44 \\ 
        ~ & \textit{GPT-4o} & 0.27 & 79.25 & 1.01 & \textbf{65.00} & 1.17 & 40.70 / 15.61 & \textbf{86.80} \\ 
        
        \bottomrule
    \end{tabular}
    }
    \caption{
    SMARTAgent's performance on the test split across three in-domain task categories. The \green{green} and \red{red} arrows indicate \green{better} or \red{worse} performance compared to the best baseline method. Its strong performance and fewer tool calls highlight SMARTAgent's efficient and strategic tool use.
    }
    \label{tab:result_main}
\end{table*}

\section{Experiment}

In this section, we present results demonstrating SMARTAgent’s effectiveness in reducing tool overuse while enhancing reasoning performance.

\subsection{Settings}


\paragraph{Data.} For in-domain testing, we evaluate SMARTAgent using the test split of adapted SMART-ER data across three domains: Math (MATH), Time (FreshQA), and Intention (IN3). For out-of-distribution (OOD) testing, we assess performance on GSM8K~\citep{cobbe2021training} and MINTQA~\citep{mintqa}, which test logical reasoning and real-world knowledge.

\paragraph{Baselines.} We incorporate three main baselines:
i) \textit{Normal Reasoning Trained}: For each domain, we train the model using the training set queries to perform reasoning without tools, leveraging the original solution chain or ground truth.
ii) \textit{Base Model Reasoning Prompt}: We directly prompt the model to apply chain-of-thought reasoning without tools to solve the problem.
iii) \textit{Base Model Tool Prompt}: We provide the model with all available tools and their usage but allow it to decide independently whether and when to use them.

\paragraph{Inference.} For reasoning without tools, the model generates a response including the final answer. For tool-reliant reasoning, the inference is interactive: in each round, if a tool call is detected, we parse and execute it, integrating the tool's output and reasoning into the input. This repeats until the final answer is reached. See \Cref{apdx:experiment} for details.

\paragraph{Metrics.} We use two main evaluation metrics: \textit{Tool Used}, which measures the average number of times a tool is leveraged during reasoning, and \textit{Accuracy}, which evaluates the average performance across queries. For the IN3 dataset, where answers depend on user preferences and lack a single correct response, we adopt the original paper’s metrics: \textit{Missing Details Recovery}, assessing whether missing details in vague instructions are recovered, and \textit{Summarized Intention Coverage}, assessing whether the final response covers all user-stated preferences.

\begin{table}[t]
    \centering

    \setlength\tabcolsep{2pt}
    \setlength\extrarowheight{2pt}
    \resizebox{1.0\linewidth}{!}{

    \begin{tabular}{l c c @{\hskip 14pt} c c}
    
        \toprule
        
        \textbf{Dataset} & \multicolumn{2}{c}{\textbf{GSM8K}} & \multicolumn{2}{c}{\textbf{MINTQA}} \\

        \cmidrule(lr){2-3} \cmidrule(lr){4-5}
        
        \textbf{Metrics} & \textbf{\makecell{Tool Used$^\downarrow$\\(\textit{Times})}} & \hskip 8pt \textbf{\makecell{Accuracy$^\uparrow$\\(\%)}} & \textbf{\makecell{Tool Used$^\downarrow$\\(\textit{Times})}} & \hskip 8pt \textbf{\makecell{Accuracy$^\uparrow$\\(\%)}} \\

        \addlinespace[2pt]
        \midrule
        \addlinespace[2pt]
        
        \multicolumn{5}{c}{ \textit{Llama-3.1-8B} }  \\ 
        \midrule

        {\fontsize{10}{10}\selectfont Normal Reasoning Trained} & 0.00 & 80.29 & 0.00 & 21.65 \\ 
        {\fontsize{10}{10}\selectfont Base Model Reasoning Prompt} & 0.00 & 82.26 & 0.00 & 12.37 \\
        {\fontsize{10}{10}\selectfont Base Model Tool Prompt} & 2.53 & 83.17 & 4.03 & 16.49 \\ 
        {\fontsize{10}{10}\selectfont \textbf{SMARTAgent}} & \textbf{0.76}$_{\green{\downarrow 1.77}}$ & \textbf{83.40}$_{\green{\uparrow 0.23}}$ & \textbf{1.06}$_{\green{\downarrow 2.97}}$ & \textbf{29.90}$_{\green{\uparrow 8.25}}$ \\
        
        \addlinespace[2pt]
        \midrule
        \addlinespace[2pt]
        
        \multicolumn{5}{c}{ \textit{Mistral-7B} } \\
        
        \midrule
        {\fontsize{10}{10}\selectfont Normal Reasoning Trained} & 0.00 & 58.68 & 0.00 & 21.65 \\ 
        {\fontsize{10}{10}\selectfont Base Model Reasoning Prompt} & 0.00 & 50.57 & 0.00 & 19.59 \\ 
        {\fontsize{10}{10}\selectfont Base Model Tool Prompt} & 3.56 & 55.34 & 6.46 & 10.31 \\ 
        {\fontsize{10}{10}\selectfont \textbf{SMARTAgent}} & \textbf{0.45}$_{\green{\downarrow 3.11}}$ & \textbf{58.98}$_{\green{\uparrow 0.30}}$ & \textbf{0.99}$_{\green{\downarrow 5.47}}$ & \textbf{25.77}$_{\green{\uparrow 4.12}}$ \\
        
        \bottomrule
    \end{tabular}
    }
    \label{tab:results_ood}
    \caption{
    SMARTAgent's performance on out-of-distribution tasks compared with baseline methods. Results show SMARTAgent can successfully generalize.
    }
\end{table}

\subsection{Main Results}
We present the main results in \Cref{tab:result_main}, along with the baseline performance of GPT-4o and GPT-4o-mini for comparison. We also present the OOD results for Mistral-7B and Llama-3.1-8B in \Cref{tab:results_ood}, highlighting the following key findings.

\paragraph{SMARTAgent solves tasks efficiently.}
Compared to the base model in \Cref{tab:result_main}, which autonomously decides whether to use tools, SMARTAgent reduces tool usage time per query by 24\% on average. At the same time, its performance improves by over 37\% across models compared to the best baseline. This demonstrates SMARTAgent's efficiency in tool use, achieving higher results while relying less on external resources.

\paragraph{7B-scale SMARTAgent can outperform GPT-4o baselines.}
Despite being much smaller, the 7B- and 8B-scale SMARTAgent models can outperform GPT-4o and its 70B counterpart in Time and Intention domains while using fewer tool calls, showcasing their efficient tool use. In Math, where reasoning scales with model size, SMARTAgent lags behind larger models but remains competitive against baselines using the same architecture. These results demonstrate that strategic tool use can bridge the gap between model size and performance, making SMARTAgent a resource-efficient yet powerful alternative.

\paragraph{SMARTAgent generalizes to OOD settings.}
As shown in \Cref{tab:results_ood}, SMARTAgent effectively reduces tool calls while achieving better overall performance on OOD test benchmarks. Notably, SMARTAgent makes only one-fifth the number of tool calls compared to the base model in MINTQA, where tool prompting often leads to excessive reliance and decreased accuracy.

\paragraph{Improper tool uses degrade performance.}  
In the MINTQA and Math domain data, we find that arbitrary tool use can degrade performance compared to standard chain-of-thought reasoning. This aligns with our argument in \Cref{sec:preliminary_llm} that excessive tool reliance can introduce unpredictable side effects, causing models to struggle with interactive tool calls. As a result, inference may become prolonged over multiple rounds, ultimately leading to incorrect answers. Additionally, we observe that larger-scale models, including GPT-4o, use tools less frequently in the Intention domain data, resulting in a greater performance drop than even the 7B-scale SMARTAgent. This may stem from their overconfidence in assisting users, leading them to overlook specific user preferences.

\paragraph{SMARTAgent achieves near-optimal tool use.}
Datasets such as Time and MINTQA contain up-to-date knowledge necessitating tool use. Ideally, at least one tool call per query is required for a correct answer, and SMARTAgent consistently maintains an average close to one, reflecting near-optimal efficiency. Similarly, in the Intention domain, where queries contain two to four missing details, SMARTAgent invokes tools three times per query, aligning with the expected need.

\begin{table*}[!t]
    \centering
    \subfigure{\includegraphics[width=\linewidth]{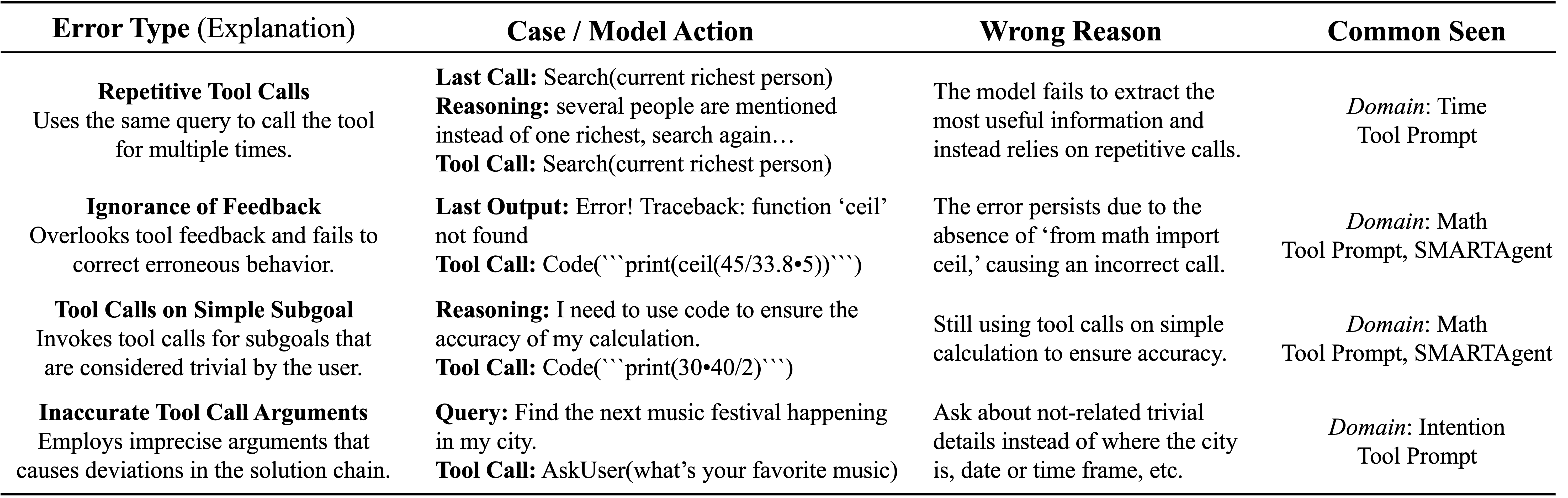}}
    \vspace{-5mm}
    \caption{Error analysis of common task failure causes, with explanations and examples.}
    \label{tab:analysis_error}
\end{table*}

\begin{table}[t]
    \centering
    
        

        


        
        
        
    \includegraphics[width=\linewidth]{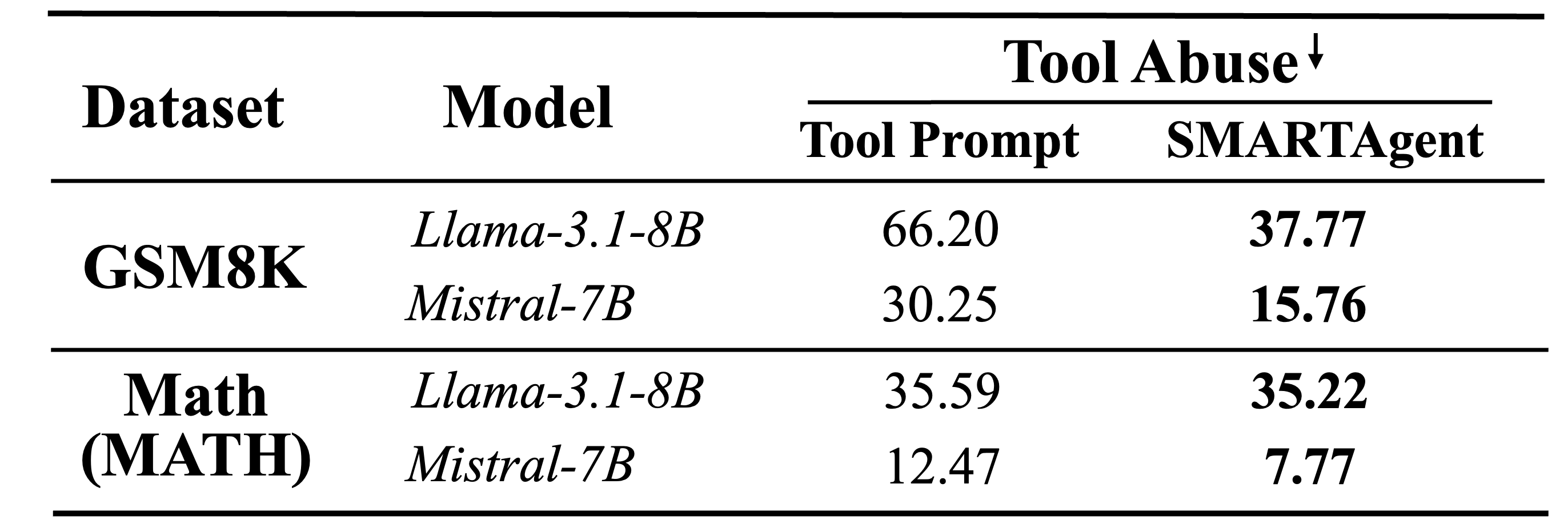}
    \vspace{-5mm}
    \label{tab:results_abuse}
    \caption{
    Statistics on tool overuse, defined in \Cref{sec:tool_overuse}.
    }
\end{table}

\begin{figure}[!t]
    \centering
    \subfigure{\includegraphics[width=\linewidth]{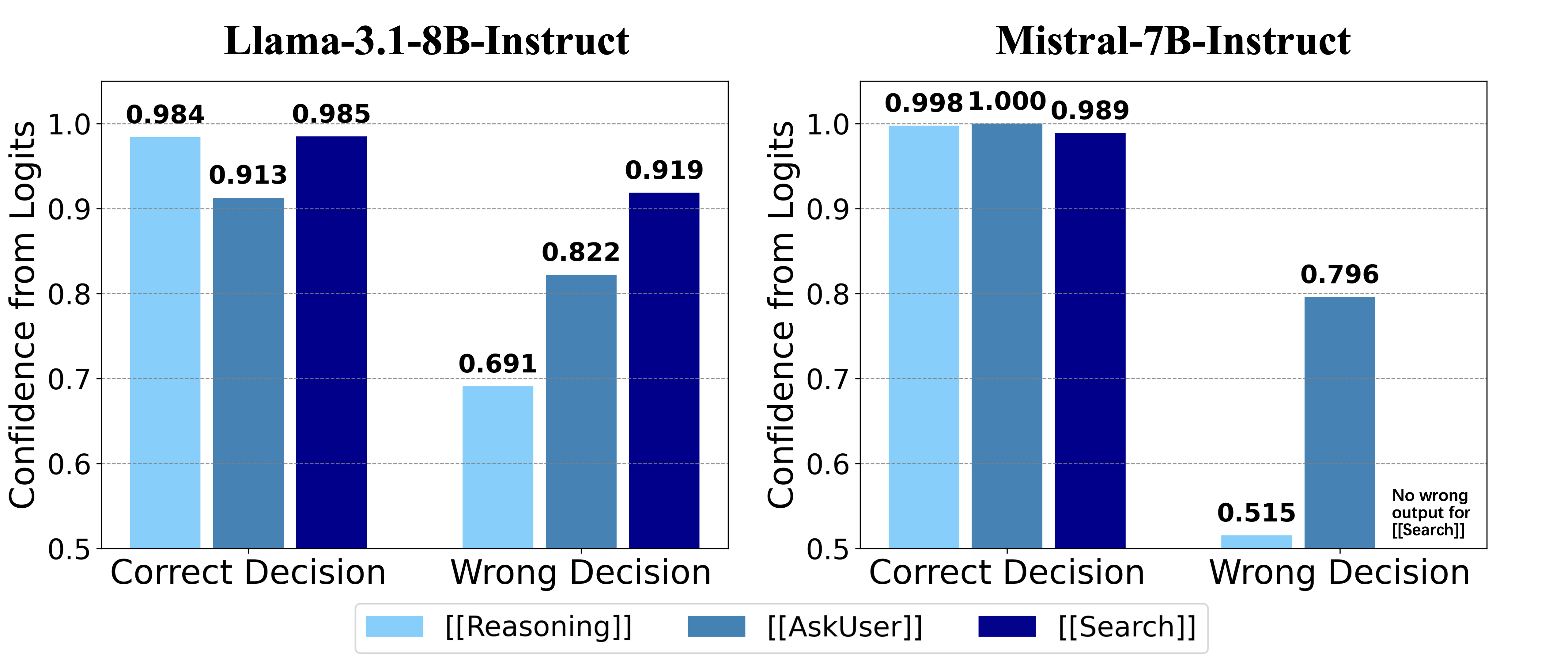}}
    \caption{Confidence analysis shows that \textbf{SMART} effectively enhances the model's decision-making confidence in selecting the correct reasoning approaches.}
    \label{fig:analysis_confidence}
\end{figure}

\begin{figure*}[!t]
    \centering
    \subfigure{\includegraphics[width=\linewidth]{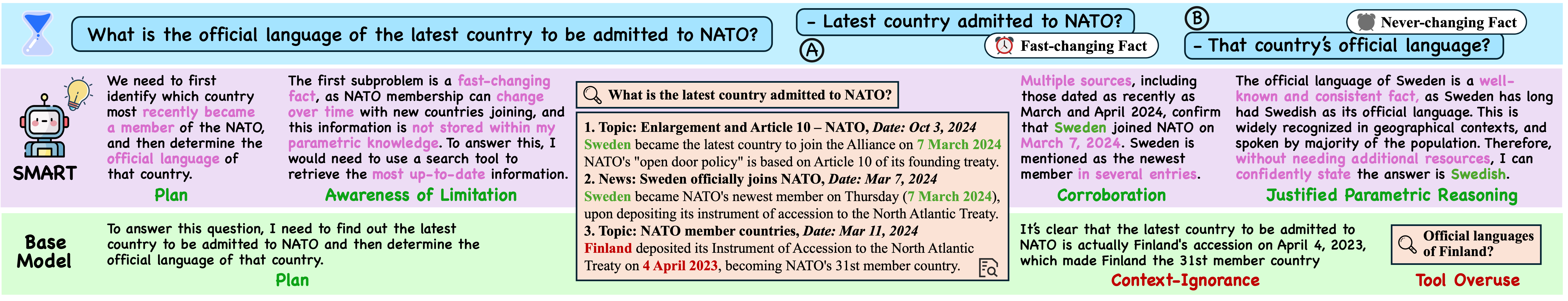}}
    \vspace{-5mm}
    \caption{Case study comparing the performance of Tool Prompting and SMARTAgent.}
    \label{fig:analysis_case}
\end{figure*}

\subsection{Analysis and Case Studies}

\paragraph{SMARTAgent effectively reduces tool overuse.} Beyond measuring tool use per query, we calculate the tool overuse rate, as defined in \Cref{sec:tool_overuse}, and report results in \Cref{tab:results_abuse} for GSM8K and Math domain test data. Notably, SMARTAgent reduces unnecessary tool calls by up to 50\% compared to prompting the base model with tool access. However, despite this reduction, tool overuse persists, which we further examine in error analysis.

\paragraph{Error analysis.} We provide error analysis in \Cref{tab:analysis_error}, highlighting common failure causes. Tool prompting leads to errors across all categories, while SMARTAgent reduces repetitive calls and improves argument accuracy. However, feedback neglect still causes tool invocation failures, particularly with the \textit{Code} tool, and excessive caution in ensuring calculation accuracy adds overhead. This mirrors human task-solving, where we sometimes rely on calculators despite knowing the steps. Future work may explore balancing convenience, budget, and efficiency to enhance decision-making.

\paragraph{Case Study.}
In \Cref{fig:analysis_case}, we compare the solution chains of SMARTAgent and the base model with tool prompting. SMARTAgent demonstrates logical planning, context corroboration, and an awareness of its limitations and knowledge boundaries, with clear justifications for its decisions. This metacognitive approach closely mirrors human reasoning processes, making SMARTAgent's reasoning more interpretable and significantly reducing tool use overhead.

\paragraph{Confidence Validation Experiment.}
To evaluate SMARTAgents' ability to choose between internal reasoning and tool invocation, we conducted experiments using special tokens to analyze decision confidence. Specifically, we trained the model on Time and Intention domains, introducing special tokens: ``[[Reasoning]]'' for internal reasoning, ``[[AskUser]]'' for the \textit{AskUser} tool, and ``[[Search]]'' for the \textit{Search} tool. These tokens, prepended at each step, guided decision-making during training (see \Cref{apdx:confidence}). For evaluation, we sampled 50 decision steps from both domains' test splits, measuring confidence via token logits. Decisions were categorized as correct or incorrect based on alignment with ground truth. As shown in \Cref{fig:analysis_confidence}, the model exhibited higher confidence in correct decisions, demonstrating \textbf{SMART's effectiveness in boosting confidence and distinguishing between internal knowledge and tool use}.

\section{Discussions}
\label{apdx:discussion}

\paragraph{Agent's improper tool usage.}
Our empirical analysis reveals a notable phenomenon of \textit{tool overuse}, where agents frequently rely on external tools even when internal knowledge is sufficient. This over-reliance likely arises from two factors: i) the agent’s uncertainty about its own capabilities, and ii) the perceived ease of external lookups compared to internal reasoning. We also observe instances of \textit{tool underuse}, especially in large-scale models like GPT-4o and Llama-70B, where agents neglect to call essential tools, possibly due to misjudging the complexity of the task. Both overuse and underuse contribute to concerns over computational efficiency and solution accuracy. Future research could explore methods to better balance these trade-offs, such as by introducing explicit resource constraints or budgets for tool calls.

\paragraph{Mechanisms behind human and LM's decision-making.}
Cognitive science suggests that human decision-making arises from both intuitive judgments and reflective strategies. Similarly, in language models (LMs), problem-solving is influenced by implicit heuristics (e.g., memorized patterns) and explicit tool-using behaviors. When tools are available, LMs often default to external queries, akin to humans seeking external confirmation when uncertain. However, unlike humans, LMs lack self-monitoring and rely on external or data-driven cues to determine when to trust their internal knowledge. Developing frameworks that integrate implicit heuristics with explicit reasoning could lead to more adaptive and efficient decision-making in LMs.

\paragraph{Enhancement of model's self-awareness.}
Our data-driven calibration strategy, which provides explicit rationales for when to rely on internal knowledge versus external tools, shows promising results. Other approaches, such as confidence probing via logits, integration of specialized self-checking modules, or reinforcement learning from feedback, might also refine tool usage thresholds. Future research could investigate how these signals affect the model's internal distributions and identify representations that capture the \textit{awareness} of boundaries. Additionally, iterative or in-context learning could allow real-time metacognitive calibration, offering a more efficient safeguard against both overuse and underuse of resources.
\section{Conclusion}
Inspired by human metacognition in decision-making, we propose the SMART paradigm for agent reasoning, where agents recognize their knowledge boundaries to decide when to use tools or parametric knowledge. Specifically, SMART-ER refines this decision boundary by incorporating questions that highlight areas where current LMs excel and struggle. Using these curated reasoning chains, we train SMARTAgent to better balance tool use and parametric knowledge, reducing tool overuse. Our results show that a simple data-driven approach can effectively calibrate model awareness, paving the way for efficient, low-resource agent development where ``smartness'' stems from both performance and metacognitive ability to optimize the reasoning strategy.


\section*{Limitations}
Our study focuses on three key domains where LLMs explicitly struggle—Math, Intention, and Time—building on insights from existing literature. However, LLMs also face challenges in areas such as long-tail knowledge and domain-specific expertise, where external resources are essential. Expanding SMART-ER to these domains could further refine model self-awareness and improve calibration in knowledge boundary, complementing the strong OOD  performance that SMARTAgent has already demonstrated. Additionally, while we evaluate our approach on two major model families, extending our analysis to a broader range of architectures, including Qwen, DeepSeek, and varying model sizes, could further validate and enhance the generalizability of our findings.

\section*{Acknowledgment}
This research is based upon work supported DARPA ITM Program No. FA8650-23-C-7316, DARPA ECOLE Program No. \#HR00112390060 and the Molecule Maker Lab Institute: an AI research institute program supported by NSF under award No. 2019897. The views and conclusions contained herein are those of the authors and should not be interpreted as necessarily representing the official policies, either expressed or implied, of the U.S. Government. The U.S. Government is authorized to reproduce and distribute reprints for governmental purposes notwithstanding any copyright annotation therein.


\bibliography{custom}

\clearpage
\appendix

\section*{Appendix}
\label{sec:appendix}

\section{Preliminary Study Details}

\subsection{Agent Experiment Details}
\label{apdx:prelim_agent}
The system instruction that we provide to both the XAgent and AgentGPT is:
{\small
\begin{tcolorbox}[colback=gray!5!white, colframe=gray!75!black, title=Prompt for Agent Preliminary Study, boxrule=0.3mm, width=0.49\textwidth, arc=3mm, auto outer arc=true]
Solve the following task accurately, and use tools to help you only if necessary.
\end{tcolorbox}
}
For LM-driven agent systems, we first prompt GPT-4o with all the questions from the GSM8K test set without using any tools. We then filter out only the questions that GPT-4o can correctly answer through pure text-based reasoning. From this refined dataset, we randomly sample 50 questions to evaluate AgentGPT and XAgent’s performance. Surprisingly, despite the core model being capable of solving all sampled questions without external tools, it still heavily relies on tools during reasoning, leading to tool overuse.

\subsection{Model Experiment Details}
\label{apdx:prelim_model}
For both Llama-3.1-8B-Instruct and Mistral-7B-Instruct-v0.3, we prompt the model to do inference two times for each question from GSM8K's test set. The first time we instruct the model to reason normally to solve the query with the following system instruction:
{\small
\begin{tcolorbox}[colback=gray!5!white, colframe=gray!75!black, title=Prompt for Model Preliminary Study (Normal), boxrule=0.3mm, width=0.49\textwidth, arc=3mm, auto outer arc=true]
You are an advanced assistant designed to solve tasks autonomously using your knowledge and reasoning. Clearly articulate your thought process and reasoning steps before presenting the final response to ensure transparency and accuracy.
\end{tcolorbox}
}

The second time, we give the model access to tools and instruct it to independently decide when to use them based on the following system instruction:
{\small
\begin{tcolorbox}[colback=gray!5!white, colframe=gray!75!black, title=Prompt for Model Preliminary Study (Tool), boxrule=0.3mm, width=0.49\textwidth, arc=3mm, auto outer arc=true]
\#\#\# Task\newline
You are a highly capable assistant designed to solve tasks effectively using your knowledge and available tools.\newline
\newline
\#\#\# Principles\newline
1. Reason Independently:\newline
• Leverage your own knowledge to analyze and solve reasoning steps whenever possible. Use external tools only when necessary.\newline
2. Tool Usage:\newline
• Use code snippet ```python ... ``` to write, execute a python code snippet, and retrieve the result from its printed output.\newline
3. Step-by-Step Approach:\newline
• Work through reasoning systematically, breaking down the task into manageable steps. Rely on your knowledge until a gap is identified that requires tool support. Employ tools to address gaps and integrate the findings into your solution.\newline
4. Goal-Oriented Resolution:\newline
• Conclude your reasoning process by achieving a clear, accurate, and succinct solution based on your independent analysis and insights gained from tools.\newline
\newline
\#\#\# Output Guidelines\newline
• If you need to use the code tool, please wrapped it ```python ... ``` and write the code snippet inside. Make sure you include all the packages necessary and the code is executable. And then you should stop generating.\newline
• If you just begin to generate reasoning steps, please directly reason after "\#\#\# Reasoning Steps".\newline
• If you are generating after the output of a code snippet, please continue to do the reasoning in your output, you can still call the tool if necessary.\newline
• Finally you should give a succinct and accurate final response to directly address the task after "\#\#\# Final Response".\newline
\end{tcolorbox}
}
We provide a code-writing and execution environment, specifically designed to assist with complex math tasks and calculations. Whenever the model generates a code snippet in its output, we parse and execute it, returning the result. The model then continues reasoning based on its previous steps and the executed output. This process iterates until a final response is reached.

\section{Data Construction Details}

\subsection{Data Selection}
\label{apdx:data_slection}
For the \textbf{Math} domain, we first collect questions that the current GPT model answers incorrectly, ensuring their inherent difficulty. We then decompose the ground truth reasoning chain to assess the complexity of each step, selecting questions that contain both straightforward and challenging aspects to provide a balanced reasoning task.

For the \textbf{Time} domain, we filter out all questions explicitly labeled as involving fast-changing facts. Given the limited number of such questions, we further augment the dataset using a self-instruct approach, prompting the GPT model to generate additional queries related to rapidly evolving information. To introduce compositional reasoning, each generated query is expanded with an additional subquestion involving well-established, slow-changing facts, forming multi-hop queries that require a nuanced understanding of temporal knowledge.

For the \textbf{Intention} domain, we filter out all queries labeled as vague in task definition, particularly those requiring explicit user clarification. To ensure that each query remains solvable without tool reliance, we probe GPT to verify that the model can generally answer each selected question without application of tools. This filtering process refines the dataset to only include queries where the model’s performance is not hindered by a lack of inherent capability but rather by the absence of user-provided intent.

The data adaptation process is fully automated, with manual checks conducted on 5\% of the samples at each stage to ensure the quality of the final filtered questions.

\subsection{Reasoning Chain Construction}
\label{apdx:chain_construction}
Empirically, we incorporate three tools in our constructed tool set:
\begin{itemize}[topsep=2pt, partopsep=-5pt, leftmargin=8pt, itemsep=-4.5pt]
\item \textbf{Code}: An environment for code writing and execution, enhancing the model’s capability in complex calculations, equation solving, and related tasks. To use this tool, the model must generate an executable code snippet within \texttt{'''python <code> '''} and print the output to obtain the execution results.
\item \textbf{Search}: A real-time web search tool for retrieving the most up-to-date factual knowledge or information beyond the model’s parametric knowledge. To invoke this tool, the model should provide a search query in the format \texttt{Search(<query>)} to obtain relevant search engine results. We empirically use the Serper API as the backend search engine.
\item \textbf{AskUser}: A tool for querying the user to clarify intentions, preferences, or general inquiries. This tool enables the model to retrieve user-provided responses by issuing a user-oriented query in the format \texttt{AskUser(<query>)}. To simulate user responses in our experiments, we employ a GPT model as the backend.
\end{itemize}
From the constructed reasoning chains, we empirically observe that the \textbf{Code} tool is mainly used in the \textbf{Math} domain, the \textbf{Search} tool is mainly utilized in the \textbf{Time} and \textbf{Intention} domains, while the \textbf{AskUser} tool is mainly employed in the \textbf{Intention} domain.

For each step involving the auxiliary model \( M \), we manually verify data quality to ensure: i) tasks are decomposed into fine-grained, reasonable subgoals, ii) tool-calling formats are correct, and iii) justifications align with labels and accurately explain why parametric knowledge suffices or a specific tool is required. Through iterative optimization of instructions to \( M \), we achieve a final pass rate exceeding 95\%.

\begin{table}[h!]
\centering
\small
\begin{tabular}{lc}
\toprule
\textbf{Hyperparameter} & \textbf{Value} \\ \midrule
Models & Llama-3.1-8B, Mistral-7B \\ 
Fine-tuning Method & SFT \\
PEFT & LoRA \\
LoRA Rank & 16 \\
LoRA Alpha & 32 \\
LoRA Dropout & 0.05 \\
LoRA Target & All Layers \\
Sequence Length (cutoff\_len) & 4096 tokens \\
Batch Size (Per Device) & 2 \\
Gradient Accumulation Steps & 4 \\
Learning Rate & 1e-4 \\
Learning Rate Scheduler & Cosine \\
Warmup Ratio & 0.1 \\
Number of Epochs & 3 \\
Precision & bfloat16 \\ \bottomrule
\end{tabular}
\caption{Hyperparameters during Fine-Tuning.}
\label{tab:hyperparams}
\end{table}

\subsection{Training}
\label{apdx:training}
For fine-tuning, we used \textbf{Llama-3.1-8B-Instruct}\footnote{\url{https://huggingface.co/meta-llama/Llama-3.1-8B-Instruct}}, \textbf{Llama-3.1-70B-Instruct}, \textbf{Mistral-7B-Instruct-v0.3}\footnote{\url{https://huggingface.co/mistralai/Mistral-7B-Instruct-v0.3}}, \textbf{Mistral-Nemo-Instruct-2407}\footnote{\url{https://huggingface.co/mistralai/Mistral-Nemo-Instruct-2407}}, and \textbf{Mistral-Small-24B-Instruct-2501}\footnote{\url{https://huggingface.co/mistralai/Mistral-Small-24B-Instruct-2501}} as base models. We applied supervised fine-tuning (SFT) in the Alpaca instruction-following format (Instruction-Input-Output), computing the loss only on tokens in the Output field.

Training was conducted on 4 NVIDIA A40 GPUs using LoRA (Low-Rank Adaptation) with a rank of 16 and an alpha of 32, applied across all model layers. The maximum sequence length was set to 4096 tokens, and models were trained for 3 epochs with a learning rate of 1e-4, using a cosine learning rate scheduler with a 10\% warmup ratio. To manage memory constraints, we set a per-device batch size of 2 and applied gradient accumulation over 4 steps. Training used \texttt{bfloat16} (bf16) precision, with evaluations every 100 steps, using 1\% of the dataset for validation. Fine-tuning hyperparameters are detailed in Table \ref{tab:hyperparams}.


The system instruction for finetuning is presented in the following:
{\small
\begin{tcolorbox}[colback=gray!5!white, colframe=gray!75!black, title=System Instruction for Training, boxrule=0.3mm, width=0.49\textwidth, arc=3mm, auto outer arc=true]
You are a highly capable assistant designed to solve tasks effectively using your knowledge and available tools. Follow these principles:\newline
\newline
1. Reason Independently: Leverage your own knowledge to analyze and solve reasoning steps whenever possible. Use external tools only when necessary.\newline
2. Tool Usage:\newline
<Specific Tool Description>\newline
3. Step-by-Step Approach:\newline
• Work through reasoning systematically, breaking down the task into manageable steps.\newline
• Rely on your knowledge until a gap is identified that requires tool support.\newline
• Employ tools to address gaps and integrate the findings into your solution.\newline
4. Goal-Oriented Resolution:\newline
Conclude your reasoning process by achieving a clear, accurate solution based on your independent analysis and insights gained from tools. After your reasoning, provide your response to directly address the task.\newline
\newline
Your reasoning should be transparent, logical, and concise. Stop and document the reasoning whenever you need to use a tool to gather more information. Continue until you reach the final solution and give final response.
\end{tcolorbox}
}

\section{Experiment Details}
\label{apdx:experiment}
\subsection{Data Setting}
\label{apdx:setting_data}
For in-domain testing, we use a subset of adapted SMART-ER data. Specifically, for the Math domain, we randomly sample 400 test instances from MATH, ensuring coverage of all testing categories (algebra, geometry, number theory, etc.), while spanning five difficulty levels. For the Time domain, we select 100 randomly sampled adapted data points from FreshQA, ensuring that each instance incorporates both fast-changing and slow-changing aspects. For the Intention domain, we randomly sample 100 data points from Intention-in-Interaction, ensuring that all selected instructions are vague and require specific user preferences to resolve.  

For out-of-domain testing, we directly use the full test set of GSM8K without modifications. For MINTQA, due to its large size, we randomly sample 10\% of the data points that meet the following criteria: the question requires multi-hop reasoning and contains both old and new knowledge. This selection ensures a challenging test set that evaluates the model’s ability to generalize beyond in-domain tasks while maintaining a focus on complex reasoning and real-world knowledge retrieval.  
\subsection{Baselines}
\label{apdx:setting_baselines}
For the baseline \textit{Normal Reasoning Trained}, we train a separate model for each domain. Specifically, for Math, Time, and Intention, we use the same queries as in the SMART-ER training set. In the Math domain, we leverage existing solution chains from the MATH dataset as training data. For the IN3 and Time domains, we use GPT-4o to generate normal reasoning chains, guided by existing annotations on final answers or missing details as heuristics. These domain-specific solution chains are then used to train the model.

For the baseline \textit{Base Model Reasoning Prompt}, we use the following system instruction to evaluate the model's performance:
{\small
\begin{tcolorbox}[colback=gray!5!white, colframe=gray!75!black, title=Base Model Reasoning Prompt, boxrule=0.3mm, width=0.49\textwidth, arc=3mm, auto outer arc=true]
You are an advanced assistant designed to solve tasks autonomously using your knowledge and reasoning. Clearly articulate your thought process and reasoning steps before presenting the final response to ensure transparency and accuracy.

In the field '\#\#\# Reasoning Steps', clearly articulate your thought process and reasoning steps towards the final answer. Then you should present a succinct and accurate final response in the field '\#\#\# Final Response'.
\end{tcolorbox}
}
For the baseline \textit{Base Model Tool Prompt}, we use the same system prompt as in \cref{apdx:prelim_model}, allowing the model to access tools and freely decide whether and when to use them.

\subsection{Interactive Inference}
\label{apdx:inference}
For both the baseline \textit{Base Model Tool Use} and our \textit{SMARTAgent}, we adopt an interactive approach for inference. Specifically, we first prompt the target model with the query and obtain its output. In this output, we use a rule-based natural language matching method to determine whether a tool call or a final answer is present (e.g., detecting whether ``\#\#\# Final Response'' appears in the output to identify the final response).  

If the final response is found, we extract it and terminate the iterative process. If a tool call is detected, we parse the parameters provided by the model to execute the tool call. Based on the specific tool's name, we invoke the corresponding API and integrate its output into the model's response. Next, we append the model's reasoning before the tool call, the tool call itself, and its output to the model's input. We then re-prompt the model to continue reasoning, given the previously executed tool call and its result.  

This iterative process continues until the final response is successfully parsed and retrieved, forming the complete interactive inference process.  

Below, we illustrate the respective input and output in an iterative inference process consisting of two iterations:
{\small
\begin{tcolorbox}[colback=gray!5!white, colframe=gray!75!black, title=Interactive Inference, boxrule=0.3mm, width=0.49\textwidth, arc=3mm, auto outer arc=true]
--------- Iterate 2 Input Begin ---------\newline
\newline
--- Iterate 1 Input Begin ---\newline
\#\#\# Task\newline
<target task>\newline
\#\#\# Reasoning Steps\newline
--- Iterate 1 Input End ---\newline
\newline
== Iterate 1 Output Begin ==\newline
- Step 1: <title>, general reasoning\newline
<reasoning>\newline
- Step 2: <title>, tool: <tool name>\newline
<tool call parameter>\newline
== Iterate 1 Output End ==\newline
- Output: <tool execution output>\newline
\newline
--------- Iterate 2 Input End ---------\newline
\newline
====== Iterate 2 Output Begin ======\newline
- Step 3: <title>, general reasoning\newline
<reasoning>\newline
- Step 4: ...\newline
...\newline
...\newline
\#\#\# Final Response\newline
<final answer>\newline
====== Iterate 2 Output End ======\newline
\end{tcolorbox}
}

\begin{table*}[t]
    \centering
    
    \setlength\tabcolsep{2pt}
    \setlength\extrarowheight{2pt}
    
    \resizebox{1.0\linewidth}{!}{
    \begin{tabular}{l @{\hskip 14pt} l c c @{\hskip 14pt} c c @{\hskip 14pt} c c c}
    
        \toprule
        
        \multirow{2}{*}{\textbf{Method}} & \multirow{2}{*}{\textbf{Model}} & \multicolumn{2}{c}{\textbf{Math} (MATH)} & \multicolumn{2}{c}{\textbf{Time} (FreshQA)} & \multicolumn{3}{c}{\textbf{Intention} (Intention-in-Interaction)} \\

        \cmidrule(lr){3-4} \cmidrule(lr){5-6} \cmidrule(lr){7-9}
        
        ~ & ~ & \textbf{\makecell{Tool Used$^\downarrow$\\(\textit{Times})}} & \hskip 8pt \textbf{\makecell{Accuracy$^\uparrow$\\(\%)}} & \textbf{\makecell{Tool Used$^\downarrow$\\(\textit{Times})}} & \hskip 8pt \textbf{\makecell{Accuracy$^\uparrow$\\(\%)}} & \textbf{\makecell{Tool Used$^\downarrow$\\(\textit{Times})}}  & \hskip 8pt \textbf{\makecell{Missing Details Recovery$^\uparrow$\\(Lv3 / Lv2, \%)}} & \textbf{\makecell{Summarized Intention \\Coverage$^\uparrow$ (\%)}}  \\

        \addlinespace[2pt]
        \midrule
        \addlinespace[2pt]
        \multicolumn{9}{c}{ \textit{Open-Source}} \\ 
        \midrule




        


        \multirow{2}{*}{\textbf{SMARTAgent}} & \textit{Llama-3.1-70B} & 0.94 & 72.50 & 1.01 & \textbf{66.00} & 3.51 & \textbf{68.60} / 58.15 & \textbf{86.09} \\
        ~ & \textit{Llama-3.3-70B} & 0.61 & \textbf{76.25} & 1.00 & 65.00 & 3.15 & 61.63 / \textbf{59.01} & 84.45 \\
        
        \bottomrule
    \end{tabular}
    }
    \label{tab:result_apdx}
    \caption{
    Performance of SMARTAgent when using Llama-3.3-70B-Instruct as the base model, compared to the original results with its Llama-3.1-70B-Instruct counterpart.
    }
\end{table*}

\subsection{Additional Results}
We also provide results from the latest \textbf{Llama-3.3-70B-Instruct}\footnote{\url{https://huggingface.co/meta-llama/Llama-3.3-70B-Instruct}} model in \Cref{tab:result_apdx}, comparing its performance with the \textbf{Llama-3.1-70B-Instruct}-based SMARTAgent. Although \textbf{Llama-3.3} is the newest version, we use the \textbf{3.1} series to maintain consistency with the \textbf{8B} model, which is also from the \textbf{3.1} version. Empirically, we found no significant difference in performance between the \textbf{3.3} and \textbf{3.1} versions of the \textbf{70B} model.

\subsection{Confidence Validation}
\label{apdx:confidence}
We independently train the Llama-3.1-8B-Instruct and Mistral-7B-Instruct models with the added special tokens. At each reasoning step, we prepend a special token at the very beginning to indicate the model’s chosen approach—whether it relies on external tools (e.g., ``[[AskUser]]'' or ``[[Search]]'') or its own parametric knowledge (e.g., ``[[Reasoning]]'').

By analyzing the probability of generating each special token, we can assess the model's confidence in its decision-making process. Apart from the added special tokens, the rest of the original reasoning chain remains unchanged, maintaining the following structured format:
{\small
\begin{tcolorbox}[colback=gray!5!white, colframe=gray!75!black, title=Step Format, boxrule=0.3mm, width=0.49\textwidth, arc=3mm, auto outer arc=true]
- Step <index>: [[Special Token]] <title>\newline
<content>\newline
- Step <index>: ...
\end{tcolorbox}
}

We train the model using the exact same hyper-parameter setting introduced in \Cref{apdx:training}. During inference, we randomly sample 50 decision-making steps from the test split of both the \textit{Time} and \textit{Intention} domains. A decision-making step refers to the final action in a reasoning sequence—given the previous \( n-1 \) steps, we evaluate whether the model correctly decides between using a tool or relying on its parametric knowledge for the \( n \)th step. This evaluation is performed within the context of the full solution chain, which consists of \( m \) steps in total (\( m \geq n \)).

\section{Additional Evaluation and Analysis}
\label{apdx:additional_exp}

To address concerns regarding dataset and model selection, we conducted additional experiments targeting two key areas: (1) the applicability of SMARTAgent on more complex reasoning benchmarks beyond GSM8K, and (2) the behavior of o1-like models with respect to tool use.

\subsection{Evaluation on Advanced Reasoning Dataset}

To assess SMARTAgent’s performance on more challenging reasoning tasks, we evaluated it on the AMC'23 benchmark~\citep{AMC2023}, a dataset known for its mathematical complexity and nuanced problem-solving requirements.

We tested two base models: Llama-3.1-8B-Instruct and Mistral-Nemo-Instruct. We compared SMARTAgent against a baseline tool prompting strategy where the tool is always made available without dynamic control.

\begin{table}[h]
\centering
\resizebox{\linewidth}{!}{
\begin{tabular}{lcc}
\toprule
\textbf{Method} & \textbf{Llama-3.1-8B-Instruct} & \textbf{Mistral-Nemo-Instruct} \\
\midrule
Tool Prompt Baseline & 12.50 & 15.00 \\
SMARTAgent & \textbf{17.50} & \textbf{20.00} \\
\bottomrule
\end{tabular}
}
\caption{Accuracy (\%) on AMC (2023) benchmark.}
\label{tab:amc}
\end{table}

\Cref{tab:amc} shows that SMARTAgent outperforms the baseline across both model backbones, highlighting its ability to handle complex tasks with improved reasoning-tool use balance.

\subsection{Behavior of o1-like Models}

To further investigate different model's tool use behavior, we conducted experiments on Deepseek-R1-Distilled variants of Llama and Qwen. Surprisingly, these models exhibited a tendency toward \emph{tool underuse}, contrary to the overuse issue our paradigm primarily targets.

\begin{table}[h]
\centering
\resizebox{\linewidth}{!}{
\begin{tabular}{lcc}
\toprule
\textbf{Model} & \textbf{Time} & \textbf{AMC (2023)} \\
\midrule
Deepseek-R1-Distilled-Llama & 60.00 & 12.50 \\
Deepseek-R1-Distilled-Qwen & 72.00 & 37.50 \\
\bottomrule
\end{tabular}
}
\caption{Tool Underuse Rate (\%) on Time and AMC tasks.}
\label{tab:underuse}
\end{table}

We define tool underuse as the rate at which a model fails to invoke a tool in scenarios where tool use would be expected.
As shown in \Cref{tab:underuse}, both distilled models significantly underutilize tools, which we attribute to potential overfitting to parametric reasoning, a phenomenon aligned with ``overthinking'' reported in prior literatures.

\subsection{SMARTAgent Adaptation for Distilled Models}

To further test the adaptability of SMARTAgent, we fine-tuned these distilled models using our SMART paradigm and evaluated them on a time-domain QA benchmark.

\begin{table}[h]
\centering
\resizebox{\linewidth}{!}{
\begin{tabular}{lcc}
\toprule
\textbf{Method} & \textbf{Distilled-Llama} & \textbf{Distilled-Qwen} \\
\midrule
Base Model Reasoning Prompt & 30.00 & 12.00 \\
Base Model Tool Prompt & 36.00 & 26.00 \\
SMARTAgent & \textbf{40.00} & \textbf{52.00} \\
\bottomrule
\end{tabular}
}
\caption{Accuracy (\%) on Time-domain QA with Distilled Models.}
\label{tab:timeqa}
\end{table}

\Cref{tab:timeqa} shows that SMARTAgent enhances both models' performance, demonstrating its effectiveness not only in mitigating overuse but also in addressing underuse by promoting strategic tool engagement. These findings reinforce SMART's broader applicability across diverse reasoning paradigms and model types.

\end{document}